\documentclass{article}

\usepackage{arxiv}

\usepackage[utf8]{inputenc} % allow utf-8 input
\usepackage[T1]{fontenc}    % use 8-bit T1 fonts
\usepackage{hyperref}       % hyperlinks
\usepackage{url}            % simple URL typesetting
\usepackage{booktabs}       % professional-quality tables
\usepackage{amsfonts}       % blackboard math symbols
\usepackage{nicefrac}       % compact symbols for 1/2, etc.
\usepackage{microtype}      % microtypography
\usepackage{lipsum}
\usepackage{graphicx}
\graphicspath{ {./images/} }

\title{Vision-based autonomous structural damage detection using data-driven methods}

\author{
 Seyyed Taghi Ataei \\
  College of Interdisciplinary Science and Technology\\
  University of Tehran\\
  Tehran, Iran \\
  \texttt{st.ataei@ut.ac.ir}  \\
   \And
 Parviz Mohammad Zadeh \\
  Faculty of New Sciences and Technologies\\
  University of Tehran\\
  Tehran, Iran\\
  \texttt{pmohammadzadeh@ut.ac.ir} \\
  \And
 Saeid Ataei \\
  Department of Systems and Enterprises\\
  Stevens Institute of Technology\\
  Hoboken, NJ 07030 \\
  \texttt{sataei@stevens.edu} \\
}

\begin{document}
\maketitle
\begin{abstract}
This study addresses the critical need for efficient and accurate damage detection in wind turbine structures, an essential component of renewable energy infrastructure. Traditional inspection methods, such as manual visual assessments and non-destructive testing (NDT), are often costly, time-intensive, and prone to human error. To overcome these limitations, this research explores the application of advanced deep learning algorithms for vision-based structural health monitoring (SHM). A dataset of wind turbine surface images, featuring categories of damage and pollution, was prepared and augmented to enhance model training. Three state-of-the-art algorithms—YOLOv7, its lightweight variant, and Faster R-CNN—were employed to detect and classify surface damage. The models were trained and tested on a dataset divided into training, testing, and evaluation subsets (80\%-10\%-10\%).

Results indicate that YOLOv7 outperformed other algorithms, achieving superior accuracy (82.4\% mAP@50) and speed, making it suitable for real-time inspections. Compared to existing studies, YOLOv7 demonstrated advancements in both detection precision and execution speed, particularly for real-time applications. However, challenges such as dataset limitations and variability in environmental conditions were noted, suggesting future work on segmentation methods and larger datasets. This research underscores the potential of vision-based deep learning techniques in transforming SHM practices by reducing inspection costs, enhancing safety, and improving reliability. These findings pave the way for more efficient and scalable monitoring solutions, contributing to the sustainable maintenance of critical infrastructures.
\end{abstract}

% keywords can be removed

\section{Introduction}

The structural health of critical infrastructures such as wind turbines plays a pivotal role in ensuring operational efficiency, longevity, and safety. Over the years, detecting and mitigating structural damage has been paramount for industries relying on these infrastructures, including renewable energy. Traditionally, structural damage detection was reliant on manual inspections, visual assessments, and non-destructive testing (NDT) methods. While these approaches offer valuable insights, they suffer from limitations such as high costs, reliance on human expertise, subjectivity, and potential safety risks for inspectors operating in hazardous conditions \cite{ref-dorafshan2017, ref-dorafshan2018}. These challenges necessitate the exploration of more efficient, reliable, and automated solutions.

Deep learning and computer vision have revolutionized the field of structural health monitoring (SHM) by automating damage detection processes and enabling high accuracy in real-time applications. Convolutional Neural Networks (CNNs), in particular, have demonstrated significant advancements in image analysis tasks, ranging from crack detection to surface anomaly classification. Their ability to autonomously learn features from large datasets has transformed traditional methods into data-driven approaches that promise improved precision, scalability, and cost-effectiveness \cite{ref-cha2017, ref-cha2018}.

This study focuses on automated damage detection in wind turbine structures using state-of-the-art deep learning algorithms. Specifically, the research investigates the performance of the YOLOv7, its lightweight variant, and Faster R-CNN, comparing their efficacy in detecting surface damage. By leveraging vision-based techniques, this work aims to address the limitations of traditional methods and highlight the potential of automated SHM systems in ensuring structural integrity.

\subsection{Importance of Automated Damage Detection}

Wind turbines are critical to the renewable energy sector, contributing to sustainable energy generation worldwide. However, they are exposed to various environmental and operational stresses, such as erosion, cracks, lightning strikes, and dirt accumulation, which can degrade their performance and shorten their lifespan \cite{ref-stock, ref-zhou}. Effective and timely damage detection is crucial for mitigating these effects, minimizing maintenance costs, and ensuring continuous energy production.

Manual inspection methods, such as climbing inspections and ground-based visual assessments, are labor-intensive and often pose safety risks. Additionally, traditional NDT techniques, including ultrasonic testing and thermography, require specialized equipment and trained personnel, making them costly and time-consuming \cite{ref-wymore, ref-jha}. In contrast, vision-based automated systems offer non-contact inspection capabilities, covering large areas efficiently while enabling real-time analysis. This shift towards automation is particularly critical for remote and offshore wind turbines, where access is often challenging \cite{ref-canizo}.

\subsection{Advancements in Deep Learning for SHM}

Recent advancements in deep learning have significantly enhanced the capabilities of image-based damage detection systems. CNNs, such as Faster R-CNN and Mask R-CNN, have been widely adopted for structural damage detection tasks, demonstrating high accuracy in detecting cracks, corrosion, and other surface anomalies \cite{ref-cha2018, ref-attard}. For instance, Faster R-CNN achieved an average mean Average Precision (mAP) of 87.8\% in detecting various damage types, including cracks and corrosion \cite{ref-cha2018}. Similarly, Mask R-CNN has been employed for crack segmentation, offering both detection and localization with high accuracy \cite{ref-attard}.

YOLO (You Only Look Once) algorithms have emerged as a breakthrough in real-time object detection, owing to their high speed and accuracy. YOLOv7, the latest iteration, introduces trainable bag-of-freebies that enhance its performance compared to earlier versions, making it highly suitable for real-time structural inspections \cite{ref-wang2022}. Studies have shown that YOLO-based models outperform traditional methods in terms of detection speed and precision, as evidenced by the application of YOLOv2 for crack detection on concrete structures \cite{ref-deng}.

\subsection{Controversial Hypotheses and Diverging Perspectives}

Despite the success of deep learning methods in SHM, some researchers argue about their scalability and reliability in real-world applications. For instance, challenges such as dataset limitations, variability in environmental conditions, and computational requirements are often cited as potential barriers \cite{ref-dorafshan2018, ref-azimi}. Additionally, while transfer learning and pre-trained models such as ImageNet have accelerated the development of SHM systems, their adaptation to specific structural contexts remains a contentious issue. Critics emphasize the need for domain-specific datasets and customized architectures to achieve optimal results \cite{ref-cui}.

Another area of debate is the trade-off between detection speed and accuracy. While YOLOv7 is celebrated for its real-time capabilities, some researchers highlight its limitations in handling small or low-contrast defects compared to multi-stage detectors like Faster R-CNN \cite{ref-shihavuddin, ref-konig}. These diverging hypotheses underscore the importance of comparative studies that evaluate multiple algorithms under diverse conditions.

\subsection{Purpose and Significance of the Study}

The primary objective of this study is to develop an automated system for detecting and localizing surface damage in wind turbine structures using image-based deep learning methods. The research aims to:

\begin{enumerate}
    \item Evaluate the performance of three deep learning algorithms—YOLOv7, tiny YOLOv7, and Faster R-CNN—in detecting and classifying surface damage.
    \item Create a comprehensive dataset of labeled images depicting various damage types, including cracks, erosion, and dirt, to facilitate supervised learning.
    \item Compare the developed system with existing studies to assess its effectiveness and highlight advancements achieved through modern algorithms.
    \item Optimize the hyperparameters of the deep learning models to improve detection accuracy and speed.
\end{enumerate}

By addressing these objectives, this research contributes to the advancement of SHM practices and demonstrates the potential of vision-based systems in reducing maintenance costs, enhancing safety, and ensuring the reliability of wind turbines.

\subsection{Overview of the Current Research Landscape}

The field of SHM has witnessed significant progress with the integration of deep learning techniques. Table \ref{tab:data_distribution} summarizes key studies and their contributions to the domain:

\begin{table}
  \caption{Key Studies in Structural Health Monitoring}
  \centering
  \begin{tabular}{cccc}
    \toprule
    \textbf{Study}                     & \textbf{Method}           & \textbf{Accuracy}      & \textbf{Weakness}                       \\
    \midrule
    Cha et al., 2018 \cite{ref-cha2018}             & Faster R-CNN        & 87.8\% mAP         & Low speed                          \\
    Dung et al., 2019 \cite{ref-dung}            & CNN                 & 90\% AP            & Not real-time                      \\
    Attard et al., 2019 \cite{ref-attard}          & Mask R-CNN          & 93.9\% accuracy    & Low speed                          \\
    Shihavuddin et al., 2019 \cite{ref-shihavuddin}     & Faster R-CNN        & 81.1\% mAP         & Low accuracy and speed             \\
    Qiu et al., 2019  \cite{ref-qiu}            & Modified YOLOv3     & 91.3\% accuracy    & Low speed                          \\
    Deng et al., 2021  \cite{ref-deng}           & YOLOv2              & 77\% mAP           & Low speed and accuracy             \\
    Ataei et al., 2025  \cite{ref-ataei}           & YOLOv7 instance segmentation              & 96.1\% mAP50         & Improved speed and accuracy        \\
    \bottomrule
  \end{tabular}
  \label{tab:data_distribution}
\end{table}

These studies highlight the effectiveness of CNN-based models in SHM while emphasizing the need for further advancements in detection speed, accuracy, and scalability. The use of innovative architectures, such as the Vision Transformer \cite{ref-dosovitskiy}, and attention mechanisms \cite{ref-cui} has shown promise in addressing these challenges.

The findings of this study indicate that YOLOv7 outperforms other algorithms in terms of accuracy and speed, making it a viable solution for real-time damage detection in wind turbines. The results also underscore the significance of dataset quality and hyperparameter optimization in achieving superior performance. By integrating vision-based deep learning methods into SHM systems, this research paves the way for more efficient, reliable, and scalable solutions in structural damage detection.

Automated damage detection systems powered by deep learning have the potential to transform SHM practices across industries. This study not only advances the state-of-the-art in wind turbine inspections but also provides a foundation for future research exploring segmentation algorithms, larger datasets, and integrated frameworks for continuous monitoring. Ultimately, these advancements contribute to the sustainable management of critical infrastructures, ensuring their longevity and operational efficiency.

\section{Research Methodology}
\label{sec:methodology}

The methodology adopted in this study focuses on developing an automated system for detecting and localizing surface damage on wind turbine structures using advanced deep learning models. This section elaborates on the dataset preparation, preprocessing techniques, model architectures, training procedures, and evaluation metrics employed in the research. By providing detailed descriptions, the methodology ensures replicability and enables future researchers to build on the findings. The primary approaches used in this study include YOLOv7, its lightweight variant (YOLOv7-tiny), and Faster R-CNN algorithms. The comparison and analysis of these methods form the cornerstone of this research. You can download the Wind Turbine Damage Dataset (Faster RCNN format) from this link: 
\url{https://www.kaggle.com/datasets/stmlen/nordtank-windturbine-dataset-faster-rcnn-format} 
and the YOLO Annotated Wind Turbine Surface Damage from: 
\url{https://www.kaggle.com/datasets/stmlen/windturbine-damage-dataset-yolo-format}.

\subsection{Dataset Preparation}

\begin{figure}
    \centering
    \includegraphics[width=0.7\linewidth]{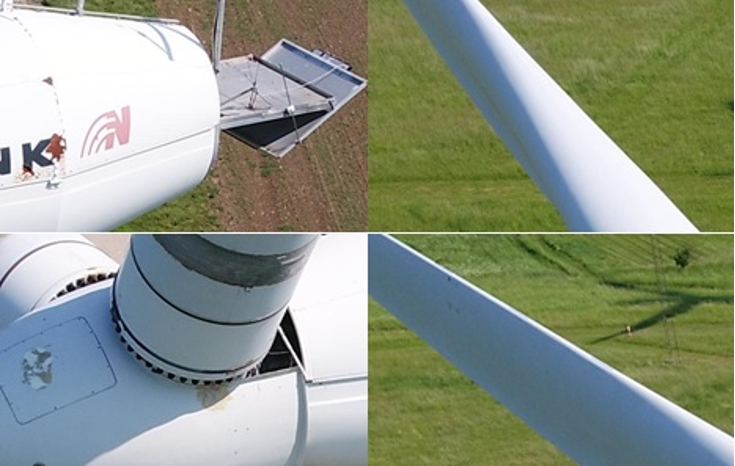}
    \caption{Sample images from the NordTank dataset \cite{ref-foster}}
    \label{fig:NordSamples}
\end{figure}

The dataset used in this study was derived from publicly available sources such as the DTU drone inspection images \cite{ref-shihavuddin2018} and the Nordtank wind turbine dataset \cite{ref-foster}. The dataset comprised 2,995 labeled images of wind turbine surfaces (some samples can be seen in Figure \ref{fig:NordSamples}, each classified into two categories: damage and pollution. These images, with a resolution of 586 x 371 pixels, were carefully selected to include diverse lighting conditions, observation angles, and damage types.

\subsubsection{Dataset Splitting}

To ensure effective training and evaluation, the dataset was divided into three subsets:

\begin{itemize}
    \item \textbf{Training Set}: 80\% of the data (2,396 images)
    \item \textbf{Testing Set}: 10\% of the data (299 images)
    \item \textbf{Evaluation Set}: 10\% of the data (300 images)
\end{itemize}

\subsection{Preprocessing Techniques}

Preprocessing was critical to standardizing the input data and enhancing model robustness. The following techniques were applied:

\subsubsection{Image Resizing and Normalization}

All images were resized to a fixed resolution suitable for the selected models. Pixel values were normalized to fall within the [0, 1] range to ensure uniformity across the dataset \cite{ref-dorafshan2018}.

\subsubsection{Data Augmentation}

To address potential overfitting and improve model generalization, data augmentation techniques were employed, including:

\begin{itemize}
    \item \textbf{Rotation and Scaling}: Simulated different viewing angles.
    \item \textbf{Flipping}: Enhanced robustness to spatial variations.
    \item \textbf{Brightness Adjustment}: Improved performance under varying lighting conditions.
\end{itemize}

\subsection{Deep Learning Model Architectures}

Three deep learning models were evaluated in this study:

\subsubsection{YOLOv7}

YOLOv7 is a one-stage object detection model known for its real-time performance and high accuracy. Its architecture includes:

\begin{itemize}
    \item \textbf{Convolutional Backbone}: Extracts hierarchical features.
    \item \textbf{Neck}: Aggregates features at different scales.
    \item \textbf{Head}: Predicts bounding boxes and class probabilities \cite{ref-wang2022}.
\end{itemize}

\subsubsection{YOLOv7-Tiny}

This lightweight version of YOLOv7 is optimized for deployment on edge devices with limited computational resources. While sacrificing some accuracy, YOLOv7-tiny maintains high detection speeds, making it ideal for real-time applications \cite{ref-wang2022}.

\subsubsection{Faster R-CNN}

Faster R-CNN is a two-stage detector that combines a region proposal network (RPN) with a CNN-based classifier. It is highly effective for detecting small or low-contrast objects but is computationally intensive compared to YOLO-based models \cite{ref-ren2015}.

\begin{figure}
    \centering
    \includegraphics[width=0.7\linewidth]{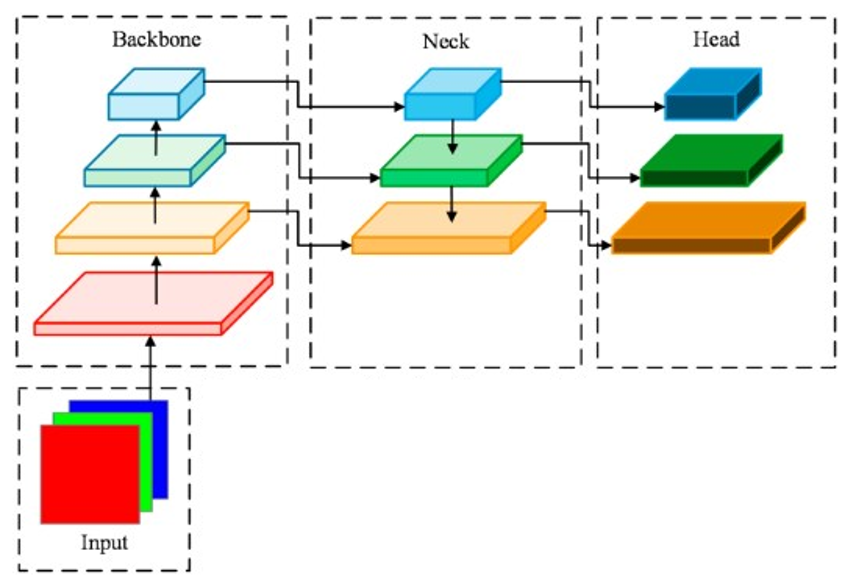}
    \caption{Architecture of YOLOv7 \cite{ref-wang2022a}}
    \label{fig:model_architectures}
\end{figure}

\begin{figure}
    \centering
    \includegraphics[width=0.6\linewidth]{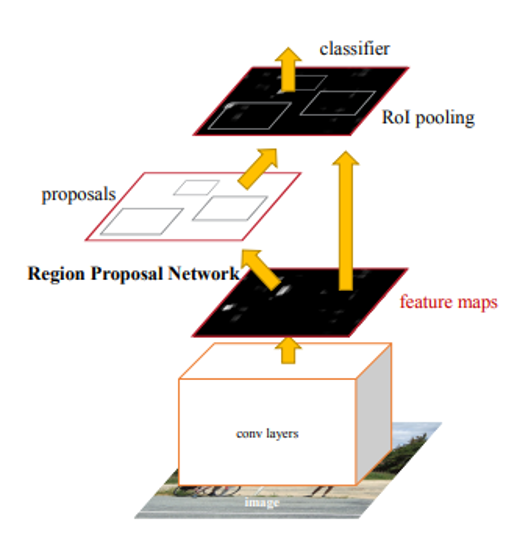}
    \caption{Architecture of Faster R-CNN \cite{ref-ren2016}}
    \label{fig:model_architecturesF}
\end{figure}

Figures \ref{fig:model_architectures} and \ref{fig:model_architecturesF} illustrate the architectures of the YOLOv7 and Faster R-CNN models.

\subsection{Training Procedures}

\subsubsection{Hyperparameter Optimization}

To optimize performance, key hyperparameters such as learning rate, batch size, and network architecture were fine-tuned for each model. The following settings were applied:

\begin{itemize}
    \item \textbf{YOLOv7}: Learning rate = 0.01; Batch size = 16; Momentum = 0.937; Epochs = 250
    \item \textbf{YOLOv7-Tiny}: Learning rate = 0.01; Batch size = 16; Momentum = 0.937; Epochs = 250
    \item \textbf{Faster R-CNN}: Learning rate = 0.001; Batch size = 64; Epochs = 10,000
\end{itemize}

\subsubsection{Transfer Learning}

Pre-trained weights were used to accelerate model convergence and improve performance. The YOLOv7 models utilized weights trained on the COCO dataset, while Faster R-CNN employed ImageNet weights for its backbone.

\subsubsection{Training Frameworks}

The models were implemented using PyTorch and TensorFlow frameworks. Training was performed on high-performance GPUs to reduce computational time. The hardware specifications included NVIDIA Tesla V100 GPUs and 32 GB RAM.

\subsection{Evaluation Metrics}

The following metrics were used to evaluate model performance:

\begin{itemize}
    \item \textbf{Precision (P)}: The proportion of correctly identified damage instances.
    \item \textbf{Recall (R)}: The proportion of actual damage instances correctly detected.
    \item \textbf{Mean Average Precision (mAP)}: Combines precision and recall for a comprehensive evaluation.
    \item \textbf{Execution Time}: Time taken to process each image.
\end{itemize}

\section{Results}
\label{sec:results}

This section presents the experimental results of using YOLOv7, YOLOv7-Tiny, and Faster R-CNN algorithms for automated damage detection in wind turbine structures. The results are divided into subheadings for clarity and include detailed metrics, performance comparisons, and observations, supported by tables and figures. These findings highlight the efficiency and applicability of deep learning models in real-world structural health monitoring (SHM).

\subsection{Model Performance Overview}

The performance of the three models was evaluated using the mean Average Precision (mAP), precision, recall, and execution time metrics. YOLOv7 demonstrated superior performance in accuracy and speed, making it the most suitable model for real-time damage detection. In contrast, YOLOv7-Tiny provided an excellent trade-off between speed and accuracy for resource-constrained environments, while Faster R-CNN excelled in detecting low-contrast and small damages but lagged in speed.

\subsubsection{YOLOv7 Performance}

\begin{figure}
    \centering
    \includegraphics[width=0.9\linewidth]{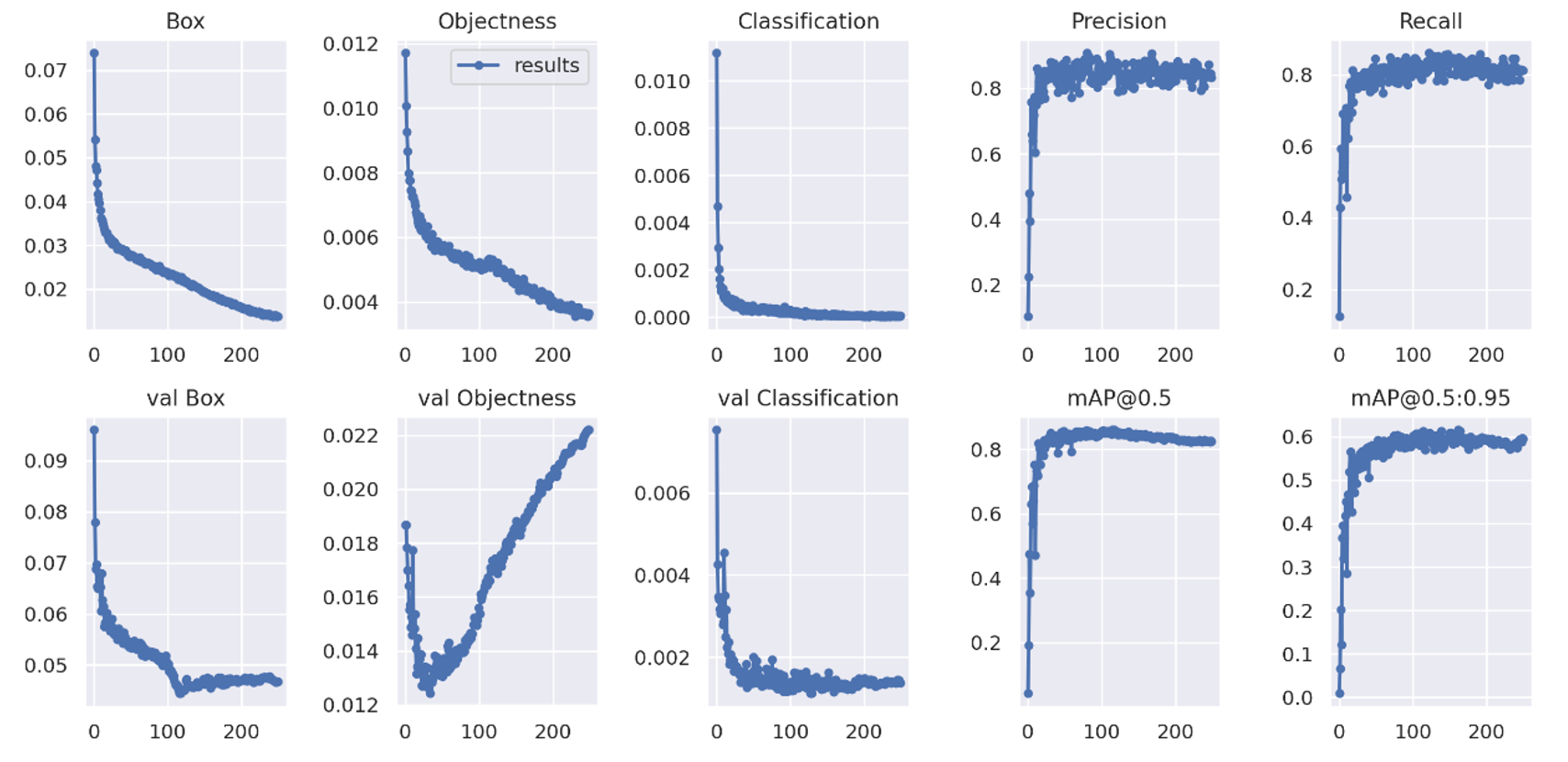}
    \caption{YOLOv7 network training results}
    \label{fig:yolov7_performance}
\end{figure}

YOLOv7 achieved an mAP@50 of 82.4\%, a precision of 83.3\%, and a recall of 81.1\%. Its average execution time was 11 ms per image, making it highly efficient for real-time applications. Figure \ref{fig:yolov7_performance} illustrates YOLOv7’s performance trends during training, showcasing its stability and high convergence rate.

\begin{table}[h]
  \caption{YOLOv7 Metrics}
  \centering
  \begin{tabular}{cc}
    \toprule
    \textbf{Metric}         & \textbf{Value} \\
    \midrule
    mAP@50                 & 82.4\% \\
    Precision               & 83.3\% \\
    Recall                  & 81.1\% \\
    Execution Time          & 11 ms \\
    \bottomrule
  \end{tabular}
  \label{tab:yolov7_metrics}
\end{table}

\subsubsection{YOLOv7-Tiny Performance}

While YOLOv7-Tiny exhibited a slightly lower accuracy (mAP@50 = 79.8\%), it excelled in speed, with an average execution time of 7 ms per image. This balance makes it ideal for edge computing and scenarios requiring lightweight models. 

\subsubsection{Faster R-CNN Performance}

Faster R-CNN achieved an mAP@50 of 79.11\% and a precision of 73.7\%, with an execution time of 200 ms per image. While its accuracy in detecting low-contrast damages was notable, the model’s slower speed limits its use in real-time applications. 

\begin{table}[h]
  \caption{Model Performance Comparison}
  \centering
  \begin{tabular}{ccccc}
    \toprule
    \textbf{Model}         & \textbf{mAP@50} & \textbf{Precision} & \textbf{Recall} & \textbf{Execution Time (ms)} \\
    \midrule
    YOLOv7                & 82.4\%          & 83.3\%            & 81.1\%         & 11                          \\
    YOLOv7-Tiny           & 79.8\%          & 80.9\%            & 77.4\%         & 7                           \\
    Faster R-CNN          & 79.11\%         & 73.7\%            & 65.6\%         & 200                         \\
    \bottomrule
  \end{tabular}
  \label{tab:model_performance}
\end{table}

\subsection{Comparative Analysis}

\subsubsection{Accuracy vs. Speed}

\begin{figure}[h]
    \centering
    \includegraphics[width=0.8\textwidth]{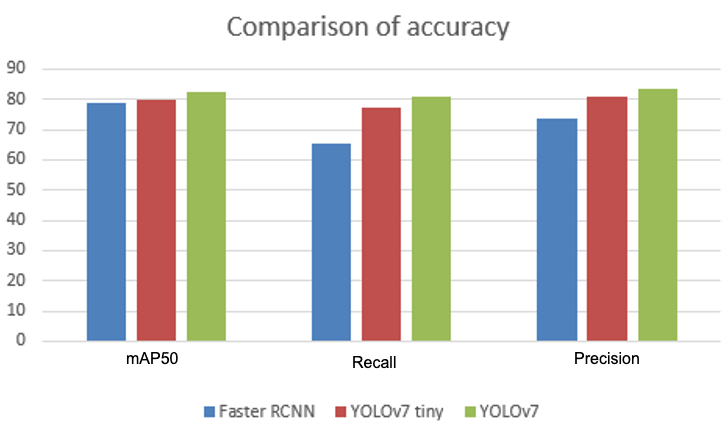}
    \caption{Accuracy of the three models}
    \label{fig:acc2}
\end{figure}

\begin{figure}[h]
    \centering
    \includegraphics[width=0.8\textwidth]{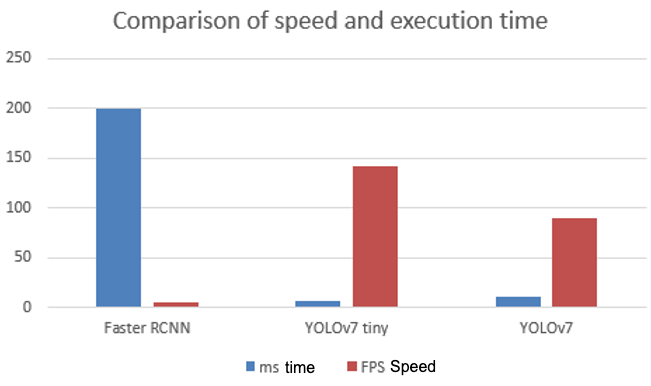}
    \caption{Speed and execution time of the three models}
    \label{fig:sp2}
\end{figure}

Figures \ref{fig:acc2} and \ref{fig:sp2} illustrate the accuracy and execution time across the three models. YOLOv7’s balance between accuracy and speed makes it the optimal choice for real-time applications. YOLOv7-Tiny, while less accurate, is highly suitable for environments with limited computational resources. Faster R-CNN, with its superior detection of low-contrast damages, remains a viable option for specialized use cases.

\subsubsection{Detection Capabilities}

YOLOv7 and YOLOv7-Tiny excelled in detecting general damage types, while Faster R-CNN was more effective in identifying low-contrast and small damages.

\begin{figure}[h]
    \centering
    \includegraphics[width=1\textwidth]{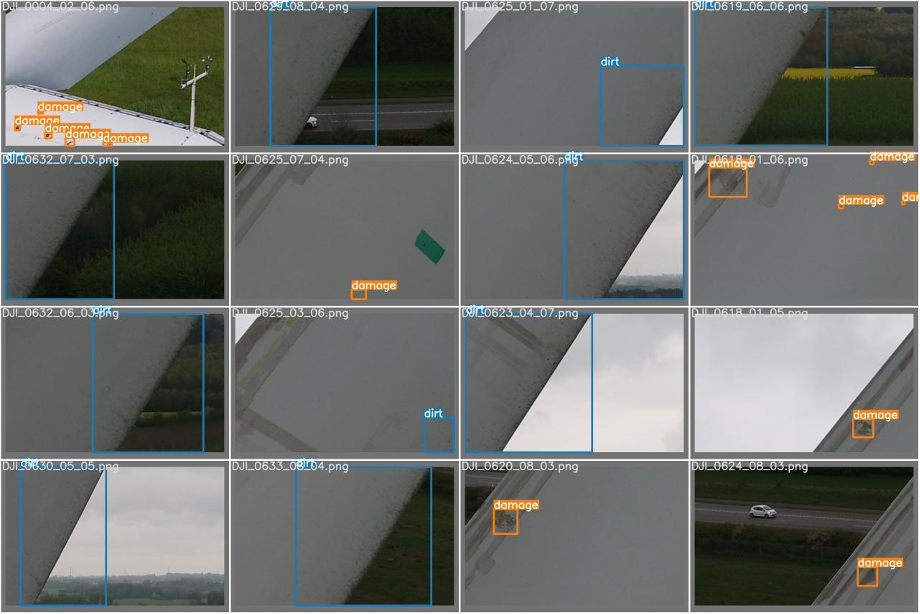}
    \caption{YOLOv7 Damage Detection Outputs}
    \label{fig:yolov7_outputs}
\end{figure}

\begin{figure}[h]
    \centering
    \includegraphics[width=0.8\textwidth]{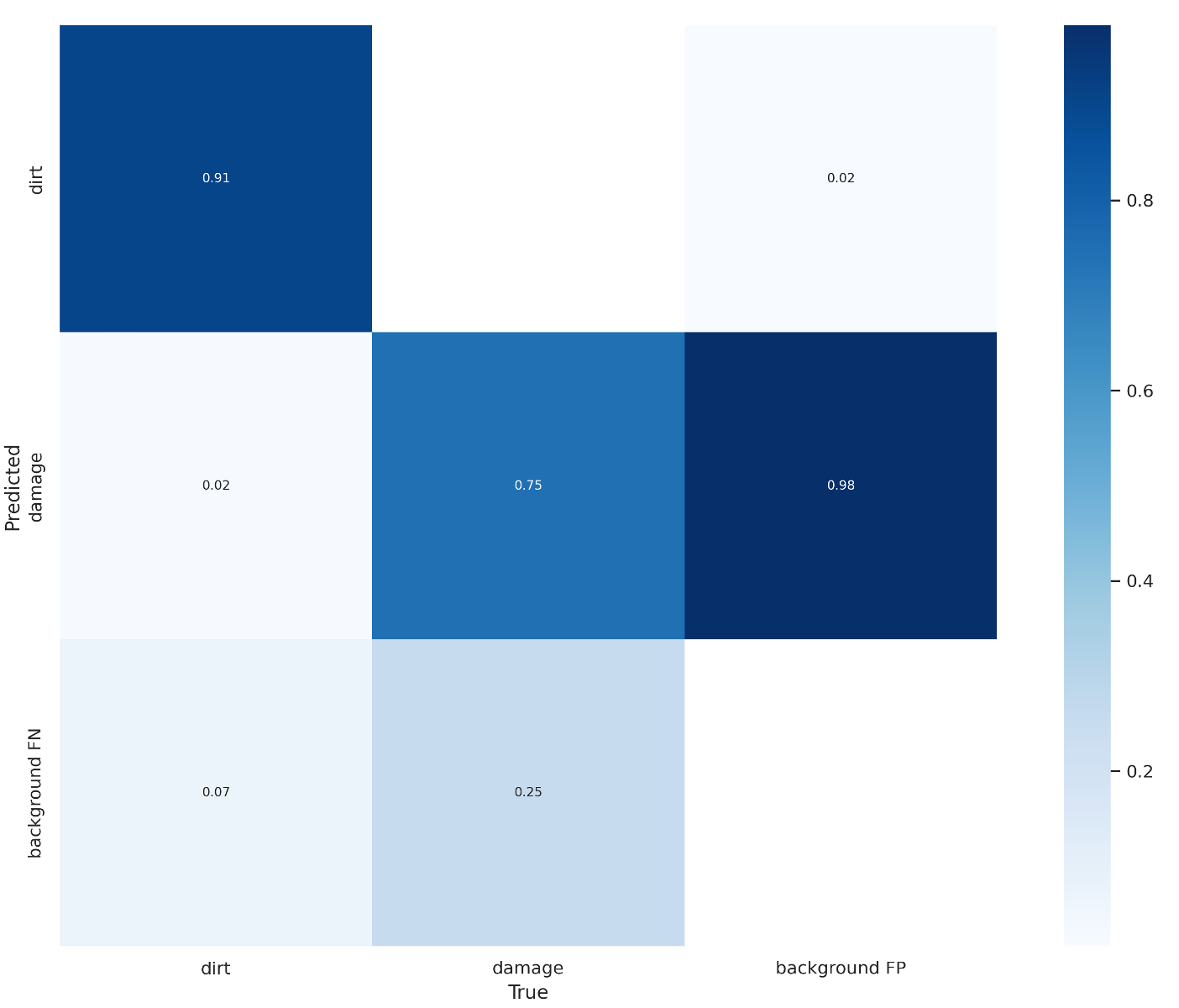}
    \caption{YOLOv7 confusion matrix}
    \label{fig:yolov7_cf}
\end{figure}

Figure \ref{fig:yolov7_outputs} showcases sample outputs from YOLOv7, highlighting its ability to accurately detect and localize surface damages under varying conditions. These visual results demonstrate the model’s robustness and applicability in diverse scenarios.
YOLOv7 confusion matrix can be seen in Figure \ref{fig:yolov7_cf}.

\subsection{Key Findings}

\begin{itemize}
    \item \textbf{YOLOv7}: Achieved the highest accuracy and speed, making it the most effective model for real-time SHM applications.
    \item \textbf{YOLOv7-Tiny}: Balanced speed and accuracy, suitable for edge computing and lightweight applications.
    \item \textbf{Faster R-CNN}: Demonstrated superior performance in detecting low-contrast damages but was limited by slower execution times.
\end{itemize}

\subsubsection{Broader Implications}

These results underscore the potential of vision-based deep learning models in transforming SHM practices. By automating damage detection, these methods reduce inspection costs, enhance safety, and improve operational efficiency. The findings also highlight the importance of selecting models based on specific application requirements, balancing accuracy, speed, and computational constraints.

\section{Discussion}

This study evaluated the efficacy of advanced deep learning algorithms—YOLOv7, YOLOv7-Tiny, and Faster R-CNN—for automated damage detection in wind turbine structures. The results revealed substantial improvements in accuracy, speed, and computational efficiency, highlighting the transformative potential of vision-based systems for structural health monitoring (SHM). This discussion examines the findings within the context of previous studies, interprets their implications, and outlines directions for future research.

\subsection{Interpretation of Results}

\subsubsection{Comparative Performance of Models}
YOLOv7 demonstrated the highest performance among the evaluated models, achieving an mAP@50 of 82.4\%, precision of 83.3\%, and an execution time of 11 ms per image. In contrast, YOLOv7-Tiny, while slightly less accurate (mAP@50 = 79.8\%), excelled in execution speed (7 ms per image), making it a viable option for resource-constrained applications. The  Faster R-CNN achieved mAP@50 of 79.11\% and a precision of 73.7\%, and an execution time of 200 ms per image, which shows its limitations in real-time applications.

The comparison in Table \ref{tab:model_performance} underscores these findings, highlighting the trade-offs between accuracy and speed across the models.

\subsubsection{Dataset Quality and Augmentation}
The quality and diversity of the dataset played a pivotal role in model performance. The dataset comprising 2,995 labeled images of wind turbine surfaces, provided a robust foundation for training. Data augmentation techniques, such as rotation, scaling, and brightness adjustment, improved model generalization.

\subsection{Broader Implications}

\subsubsection{Advancing SHM Practices}
The adoption of YOLOv7 and its variants in SHM systems marks a significant advancement in automated inspection methodologies. Traditional inspection methods, such as manual assessments and non-destructive testing (NDT), are often labor-intensive and time-consuming \cite{ref-wymore}. By contrast, vision-based systems offer rapid, accurate, and non-contact solutions, reducing maintenance costs and enhancing safety. 

\subsubsection{Real-Time Applications}
The execution speed of YOLOv7 and YOLOv7-Tiny enables their deployment in real-time applications, such as on-site inspections and integration with drones. This capability is particularly valuable for offshore wind turbines, where accessibility is limited \cite{ref-canizo}. The use of YOLOv7-Tiny for edge computing further extends its applicability to scenarios requiring lightweight and portable solutions.

\subsection{Challenges and Limitations}

\subsubsection{Computational Requirements}
Although YOLOv7 and its variants demonstrated high efficiency, the training process remains computationally intensive, requiring access to high-performance GPUs. This limitation may hinder adoption in resource-constrained settings.

\subsubsection{Dataset Constraints}
Despite the robustness of the dataset, its limited representation of certain damage types and environmental conditions highlights the need for larger, more diverse datasets. Expanding the dataset to include more complex scenarios, such as varying weather conditions, could further enhance model robustness.

\subsection{Future Research Directions}

\subsubsection{Dataset Expansion}
Future research should focus on expanding datasets to include diverse damage types, environmental conditions, and real-world noise. Integrating drone-captured images, as proposed by Shihavuddin et al. (2019), could further enhance dataset diversity.

\subsubsection{Incorporation of Segmentation Algorithms}
Future work could explore the use of segmentation models, such as Mask R-CNN or U-Net, to provide more detailed insights into damage extent and severity.

\subsubsection{Deployment on Edge Devices}
Optimizing YOLOv7-Tiny for edge devices could enable real-time inspections in resource-constrained environments.

\section{Conclusions}

This study evaluated the performance of advanced deep learning algorithms—YOLOv7, YOLOv7-Tiny, and Faster R-CNN—for automated damage detection in wind turbine structures. The findings underscore the transformative potential of vision-based methods in improving structural health monitoring (SHM) efficiency and reliability. Each model’s strengths and limitations provide insights into their optimal applications, contributing to a robust foundation for future research and practical implementations.

YOLOv7 emerged as the most effective model for real-time damage detection. It achieved an mAP@50 of 82.4\%, precision of 83.3\%, and recall of 81.1\%, with an average execution time of 11 ms per image. This balance of accuracy and speed makes YOLOv7 particularly suitable for real-time SHM applications. 

YOLOv7-Tiny offers a practical solution for resource-constrained environments. While its accuracy (mAP@50 = 79.8\%) is slightly lower than YOLOv7, its faster execution time of 7 ms per image makes it ideal for edge computing and lightweight applications. This trade-off supports its deployment in scenarios requiring high efficiency.

Faster R-CNN demonstrated notable performance in detecting low-contrast and small damages, achieving an mAP@50 of 79.11\% and precision of 73.7\%. However, its slower execution time (200 ms per image) limits its applicability in real-time scenarios. 

While challenges such as low-contrast damage detection and computational requirements persist, the integration of diverse datasets, hybrid architectures, and segmentation algorithms offers promising avenues for future research. By advancing SHM practices, this research contributes to the sustainable maintenance of critical infrastructures.

\bibliographystyle{unsrt}  


\end{document}